\pdfoutput=1

\documentclass[11pt]{article}

\usepackage{ACL2023}

\usepackage{times}
\usepackage{latexsym}

\usepackage[T1]{fontenc}

\usepackage{microtype}

\usepackage{inconsolata}

\usepackage[utf8]{inputenc}
\usepackage{booktabs}
\usepackage{array}\newcolumntype{P}[1]{>{\centering\arraybackslash}p{#1}}
\usepackage{makecell}
\usepackage{enumitem}
\usepackage{graphicx}
\usepackage{multicol}
\usepackage{multirow}
\usepackage{stfloats}
\usepackage{float}
\usepackage{amsmath}
\usepackage{bm}

\fnbelowfloat

\usepackage{microtype}

\usepackage{hyperref}

\usepackage{todonotes} 

\title{NLP Reproducibility For All: Understanding Experiences of Beginners}

\author{Shane Storks  \hspace{30pt} Keunwoo Peter Yu  \hspace{30pt} Ziqiao Ma \hspace{30pt} Joyce Chai \\
  Computer Science and Engineering Division, University of Michigan \\
  \texttt{\{sstorks, kpyu, marstin, chaijy\}@umich.edu} \\
}

\begin{document}

\maketitle

\begin{abstract}

As natural language processing (NLP) has recently seen an unprecedented level of excitement, and more people are eager to enter the field, it is unclear whether current research reproducibility efforts are sufficient for this group of \textit{beginners} to apply the latest developments.
To understand their needs, we conducted a study with 93 students in an introductory NLP course, where students reproduced the results of recent NLP papers. Surprisingly, we find that their programming skill and comprehension of research papers have a limited impact on their effort spent completing the exercise. Instead, we find accessibility efforts by research authors to be the key to success, including complete documentation, better coding practice, and easier access to data files. Going forward, we recommend that NLP researchers pay close attention to these simple aspects of open-sourcing their work, and use insights from beginners' feedback to provide actionable ideas on how to better support them.

\end{abstract}

\section{Introduction}

As natural language processing (NLP) research continues to grab public attention {and excitement}, it becomes increasingly important for it to be accessible to a broad audience. While the research community works to democratize NLP, it remains unclear whether \textit{beginners} in the field can easily apply the latest developments.
How easy is it for them to reproduce experimental results? Will their programming background or comprehension of papers play a role? What key elements affect their experience here?

To address these questions, we conducted a {controlled user study} in an introductory NLP class. {We first identified and successfully reproduced the results of three recent NLP publications ourselves. We surveyed students on their background in machine learning and programming, then divided them into three skill level groups based on the survey results.}
Each group was asked to reproduce the results of the three papers. As students conducted experiments, they tracked their own efforts, while their computational resource usage was logged by a shared high-performance computing system.
After conducting the experiments, students answered questions and provided feedback about their experiences.

Our results show that beginners' technical skill level and comprehension of the papers play only a small part in their experience in reproducing the results, and we observed a strikingly wide range of time spent on the exercise regardless of these user-specific factors.
{Meanwhile, we show that reproducibility efforts by paper authors make a much larger impact on user experience, and
based on direct feedback from these beginners, we find that they encounter a range of roadblocks related to the documentation and ease of use of open-sourced materials.}
These findings shed light on the extra steps NLP researchers can take to make state-of-the-art technology more accessible to beginners, an important direction to continue democratizing NLP to its growing audience. {To this end, we make several concrete recommendations in Section~\ref{sec:discussion} on how the NLP research community can further improve accessibility to beginners.}

\section{Related Work}

Amidst growing concern of a reproducibility crisis across many scientific disciplines~\cite{baker2016}, recent years have seen an increasing amount of work studying trends and proposing guidelines to improve research reproducibility in NLP and neighboring disciplines \cite{arabas2014case,rozier2014reproducibility,cvrepro,Henderson_Islam_Bachman_Pineau_Precup_Meger_2018,crane-2018-questionable,cohen-etal-2018-three,tatman-2018-taxonomy,dodge-etal-2019-show,mtchecklist,pineau2021improving,rogers-etal-2021-just-think,belz-2022-metrological}.
As part of these efforts, the Association for Computational Linguistics (ACL) has also adopted author checklists for reproducibility and responsible research practice.

As a prerequisite for reproducibility in NLP, prior work has studied the availability of code and data at scale \cite{wielingCL}.
Where code and data are available, large-scale and multi-test studies have assessed the reproducibility of results \cite{antonio-rodrigues-etal-2020-reproduction,branco-etal-2020-shared}.
Various venues have invited individual reproductions and replications of work in machine learning and NLP, including the
Reproducibility, Inexplicability, and Generalizability of Results 
Workshop~\cite{arguello2015} in information retrieval, the Reproducibility in ML workshops\footnote{\url{https://sites.google.com/view/icml-reproducibility-workshop/}} at the International Conference on Machine Learning (ICML) and International Conference on Learning Representations (ICLR), and the ML Reproducibility Challenges\footnote{\url{https://paperswithcode.com/rc2021}} at ICLR and Neural Information Processing Systems (NeurIPS). \citet{belz-etal-2021-systematic} reviewed and aggregated many of these efforts in NLP, finding only up to
14\% of experiments led to accurately reproduced results.
Unlike prior work, we are not focused on the accuracy of results, but rather NLP beginners' experiences in reproducing results that are pre-verified to be reproducible.

\section{Methodology}
Next, we introduce the methodology for our study, including the steps taken for data collection, and the variables of interest we study in our analysis of beginners' experience reproducing NLP results.

\subsection{Data Collection}
Data collection for this study consisted of several steps, outlined below.

\subsubsection{Pre-Survey}\label{sec:skill level}
First, we conducted a pre-survey on students' backgrounds and their understanding of course material. We pose our questions based on the categories proposed by \citet{feigenspan2012measuring} to measure programming experience, and ask students to provide informed consent to use their survey responses for the purpose of this study.
The full set of pre-survey questions 
can be found in Appendix~\ref{apx: presurvey}.

\subsubsection{Paper Selection \& Expert Reproduction}\label{sec: selection}
{Next, we carefully selected a small number of papers in recent ACL conferences, verifying that we could reproduce their results in a comparable, reasonable amount of time and effort using a single GPU\footnote{This restriction on effort was necessary as work was completed in a university course, but it also may have reduced the impact that paper-specific roadblocks could have on time taken to reproduce results, and ultimately students' outlooks.} from an institution-provided computing cluster managed with Slurm.\footnote{\url{https://slurm.schedmd.com/}} In line with the definition of reproducibility from \citet{rougier2017sustainable}, we used the code and data published by the paper authors to attempt to reproduce results, rather than re-implementing models and algorithms.
Out of 24 papers considered from 2018 through 2022, we could accurately reproduce results within these bounds from only 3 papers.
Common reasons for failure to reproduce results included long model training time requirements, incompatibility of code bases with our computing platform, incomplete documentation, and discrepancies between reproduced results and those reported in papers.
Selected reproducible papers are identified in Table~\ref{tab: paper info},\footnote{{Papers identified at reviewers' request.}} while the selection process is fully documented in Appendix~\ref{apx: paper selection}. Selected papers come from one track (Sentence-Level Semantics and Textual Inference), minimizing the impact of varying subject matter complexity among different topic areas.}

It is important to note that the goal of this research is toward user reproducibility \textit{experience}, which is different from previous works focusing on the accurate reproducibility of results. Therefore, instead of having many papers to conduct the experiments, we chose to control the study by selecting a small number of comparable papers {with different characteristics in their released code}. As each paper was reproduced by a group of students with different skill levels, we expected to gather sufficient statistics to identify common trends in responses to these characteristics.

\subsubsection{Reproduction Process}\label{sec:reproducibility study}
In reproducing results themselves, students were required to use
the same GPU computing resources {within our university's shared cluster} to control for the potential effect of computing infrastructure on results. {Students used Slurm to request and acquire sole control of resources,} and all resource utilization was automatically tracked by this centralized platform. 
To further control the impact of our specific computing environment, we ensured students' familiarity with the platform through an earlier assignment to implement and run state-of-the-art NLP models on the platform.
Each student was assigned to reproduce results from one of the selected papers (A, B, or C); more information on paper assignment is provided in Section~\ref{sec: skill level assignment}.
While reproducing experiments from their assigned paper, students tracked time spent setting up the code base, {as directed in a homework assignment associated with this study.}\footnote{{Instructions given to students are listed in Appendix~\ref{apx: hw content}.}}

\subsubsection{Post-Survey}\label{sec:post-survey}
After reproducing the results of each paper, students completed a survey about their experience. Here, we asked questions about their comprehension of the paper, time spent and difficulties encountered reproducing the results, and general outlooks on the experiment, including what helped and blocked their ability to reproduce results.
Post-survey questions are listed in Appendix~\ref{apx: postsurvey}.

\subsection{Analysis of User Experience}
Next, we introduce the key factors in interpreting our collected data, and how they may be characterized.
Specifically, we consider a student's experience reproducing the results of their assigned paper as a dependent variable, and consider three types of independent variables: students' skill level, students' comprehension of a paper, and paper authors' specific efforts toward making results reproducible. Our goal is to understand how each of these variables can impact a beginner's experience in reproducing NLP results.

\subsubsection{Defining User Reproducibility Experience}\label{sec:setup time runtime}
In this reproducibility study, we aim to understand a beginner's \textit{experience} in reproducing results. We characterize this by students' \textbf{time spent} and \textbf{reported difficulty} to reproduce results.

\paragraph{Setup time and runtime.}
Time spent to reproduce results is divided into two phases:
\vspace{5pt}
\begin{enumerate}[noitemsep, nolistsep]
    \item \textbf{Setup time:} Downloading and setting up the code, dataset, and external dependencies.
    \item \textbf{Runtime:} Training and evaluating systems.
\end{enumerate}
\vspace{5pt}

System setup time is self-reported {in the post-survey}, while runtime is extracted from the centralized {Slurm system using its resource usage tracking feature, which accurately reports the total GPU usage time per student. As such, runtime may include extra time a student may have spent for trial and error, including requesting extra GPUs that went unused. Calculating runtime this way is suitable for our purpose, as management of hardware resources could pose a real barrier to NLP beginners.}
As runtime varies significantly by paper, we quantify it by percent error from the research team's runtime when reproducing the same result. These variables can provide indirect yet objective measures on how much a student struggled with the experiment compared to other beginners and experts.

\paragraph{Difficulty ratings.}
For a more direct measure of students' experience, we also considered student ratings for difficulty encountered in each step of the experiment (on a scale from 1-5, 5 being most difficult):
\begin{enumerate}[noitemsep, nolistsep]
    \item \textbf{Downloading source code}, which requires cloning one or more GitHub repositories.
    \item \textbf{Downloading data}, which may be hosted somewhere different than the code.
    \item \textbf{Setting up the code base}, including installing external dependencies or pre-trained models.
    \item \textbf{Data preprocessing}, which may entail running scripts or manual adjustment.
    \item \textbf{System training}, which may require hyperparameter search or be informed by a pre-selected hyperparameter configuration.
    \item \textbf{System evaluation}, where evaluation metrics directly comparable to the paper's must be calculated and reported.
\end{enumerate}

\begin{table}
    \centering
    \footnotesize
    \begin{tabular}{P{1.0cm}P{2.8cm}P{1.0cm}P{1.1cm}}\toprule
        \textbf{Paper} & {\textbf{Reference}} & \textbf{{Setup}} & \textbf{{Runtime}} \\\midrule
        A & \citet{zhou-etal-2021-temporal} & 2 hrs. & 0.5 hr. \\
        B & \citet{donatelli-etal-2021-aligning} & 2 hrs. & 3 hrs. \\
        C & \citet{gupta-etal-2020-infotabs} & 2 hrs. & 2 hrs. \\\bottomrule
    \end{tabular}
    \normalsize
    \caption{Selected papers for the study, and research team's code setup time and runtime\footnotemark \hspace{1pt} rounded to the nearest half hour. All papers are from the \textit{Sentence-Level Semantics and Textual Inference} area in ACL venues.}
    \label{tab: paper info}
    \vspace{-15pt}
\end{table}
\footnotetext{Calculation of setup time and runtime described in Section~\ref{sec:setup time runtime}.}

\subsubsection{Defining Skill Level}\label{sec: skill level assignment}
We may expect that a student's skill level or technical background may have an impact on their experience. 
As such, we collected data about students' programming background and understanding of NLP coursework in the pre-survey. To characterize student skill level, four variables are extracted from their responses:
\begin{enumerate}[noitemsep, nolistsep]
    \item \textbf{Python experience} (years)
    \item \textbf{PyTorch experience} (years)
    \item \textbf{LSTM understanding} (1-5 from worst to best understanding)
    \item \textbf{Transformer understanding} (1-5 from worst to best understanding)
\end{enumerate}

All skill level factors are self-reported. 
We focus on Python and PyTorch as these are commonly used in NLP research, including all selected papers.
As such, knowledge of them may most directly transfer to reproducing NLP results. 
Meanwhile, students' understanding of LSTMs~\cite{hochreiter1997long} and transformers~\cite{vaswani2017attention} is self-reported in the pre-survey based on 
related past homework assignments {requiring students to implement them in PyTorch. This hands-on experience could contribute to their ability to reproduce results from the selected papers, each of which applied transformer-based models.}
Accounting for these factors equally, we divide the 93 study participants into 3 skill level groups as close to equal size as possible (considering ties): {\bf \em novice}, {\bf \em intermediate}, and {\bf \em advanced}.
As shown in Table~\ref{tab: paper assignments}, papers were distributed mostly uniformly within each skill level.\footnote{{Subject consent caused minor non-uniformity in assignment distributions.}}

\begin{table}
    \centering
    
    \begin{tabular}{P{1.5cm}P{1.0cm}P{1.0cm}P{1.0cm}P{1.0cm}}\toprule
        \textbf{Paper} & \textbf{Nov.} & \textbf{Int.} & \textbf{Adv.} & \textbf{Total} \\
        \cmidrule(lr){1-1} \cmidrule(lr){2-4} \cmidrule(lr){5-5}
        A & 12 & 11 & 11 & 34 \\
        B & 10 & 10 & 10 & 30 \\ 
        C & 10 & 9 & 10 & 29 \\\bottomrule
    \end{tabular}
    \normalsize
    \caption{Distribution of assigned papers across skill level groups (novice, intermediate, and advanced).}
    \vspace{-5pt}
    \label{tab: paper assignments}
\end{table}

\subsubsection{Defining Comprehension}\label{sec: define comprehension}
Meanwhile, we might expect that a student's comprehension of a specific paper could also contribute to their ability to reproduce its results.
To measure students' comprehension of a paper objectively, we carefully designed a set of four-way multiple-choice questions about the key aspects of each work. Specifically, we asked about each paper's:

\begin{enumerate}[noitemsep, nolistsep]
    \item \textbf{Motivation}: Prior limitations in related work addressed by the paper.
    \item \textbf{Problem Definition}: Target task details.
    \item \textbf{Approaches}: Inputs and outputs of reproduced system.
    \item \textbf{Implementation}: Matching of a process described in the paper to a file in the code.
    \item \textbf{Results}: Evaluation criteria details.
    \item \textbf{Conclusion}: Implications of results.
\end{enumerate}

Students answered these questions in the post-survey.
Questions were posed in such a way that the answers could not be found directly in the paper or code. While the specific question and answers of course varied by paper, their nature was standard, enabling us to consistently characterize students' comprehension of a paper.
Together, answering these questions correctly implies a comprehensive understanding of the work which may be supported not just by reading the paper, but also working hands-on with the code. As such, we measure comprehension of the paper by students' accuracy on these questions as a whole, and can even use their correctness on specific questions to represent comprehension of specific aspects. {The specific comprehension questions we asked students are listed in Appendix~\ref{apx: comprehension questions}.}

\subsubsection{Defining Author Reproducibility Efforts}\label{sec: define repro effort}
A strong source of guidance for reproducibility in NLP is the ACL Reproducibility Checklist (ACLRC)\footnote{See Appendix~\ref{apx: acl checklist text} for the full checklist.} which authors must complete in order to submit manuscripts to many ACL publication venues.\footnote{More recently, some items of this checklist have been adapted into the more comprehensive Responsible NLP Research checklist used at some venues \cite{rogers-etal-2021-just-think}. We compare these checklists in Section~\ref{sec:discussion}.}
While the items listed on the ACLRC are all important efforts for the reproducibility of NLP research in general, we would like to understand which of them are particularly important for beginners' success.

This is not straightforward for a few reasons. 
First, whether an item on the checklist is satisfied is subjective, as a paper's code release may provide some information pertaining to an item, but the degree to which it is easy to find and understand may vary. 
Second, the reproducibility of some papers may benefit from efforts toward certain items more than others. For example, if a paper entails a long model training time, reporting the expected training time may be especially beneficial for users of pay-per-use computing environments. 
Lastly, the degree to which a reproducibility effort is found helpful by users can be subjective. For example, one may find a reproducibility effort helpful just by its existence regardless of its quality, e.g., an unclear hyperparameter configuration buried in the code, while others may prioritize high quality.

For the most complete understanding of beginners' experiences, we make no restrictions along these lines. For each paper, students selected items from the ACLRC that they specifically found to be most helpful toward reproducing the results of their assigned paper.\footnote{Summary of student responses in Appendix~\ref{apx: checklist items}.} We use their responses to characterize each paper's efforts toward reproducibility.

\section{Results \& Findings}
Here, we investigate how aspects of students' background and understanding of papers, as well as reproducibility efforts by the authors of papers, affected students' experience in reproducing experimental results.\footnote{Data analysis code shared at \url{https://github.com/sled-group/NLP-Reproducibility-For-Beginners}.}
Following prior work, we first verify the accuracy of the results. Students on average produced results with a relative error within 3\% accuracy from reported results regardless of their skill level or assigned paper.\footnote{See Appendix~\ref{apx: relative error} for more on accuracy of results.} 
This is expected, as we verified that all papers' results can be accurately reproduced.

Next, we systematically investigate the experience students had reproducing results from the papers. We analyze their experience from three perspectives: students' skill level (Section~\ref{sec: skill level assignment}), students' comprehension of their assigned paper (Section~\ref{sec: define comprehension}), and paper authors' efforts toward making their results more reproducible (Section~\ref{sec: define repro effort}).

\begin{table}
    \centering
    \footnotesize
    \begin{tabular}{lcccc}\toprule
        \textbf{Skill Level Factor} & \textbf{$\boldsymbol{\rho}$ (time)} & \textbf{$\boldsymbol{\rho}$ (diff.)} \\
        \cmidrule(lr){1-1} \cmidrule(lr){2-3} 
        Python Experience (Years) & -0.291 & -0.230 \\
        PyTorch Experience (Years) & -0.251 & -0.259 \\
        LSTM Understanding (1-5) & \textbf{-0.430} & \textbf{-0.396} \\
        Transformer Understanding (1-5) & -0.317 & -0.338 \\
        \bottomrule
        
    \end{tabular}
    \normalsize
    \caption{Spearman correlation coefficients for how well self-reported code setup time and difficulty is predicted by skill level factors, including PyTorch and PyTorch experience, and self-reported understanding of NLP models covered in course homework assignments. All correlations are statistically significant ($p<0.05$); strongest correlations for each dependent variable in bold.}
    \label{tab: skill level vs code setup}
    \vspace{-10pt}
\end{table}

\subsection{Student Skill Level}
First, we examine the relationship of a student's skill level with their experience, specifically the time taken to set up and run their experiment, as well as their opinion on how difficult it was.

\begin{figure}
    \centering
    \hspace{-0.5em}\includegraphics[width=0.23\textwidth]{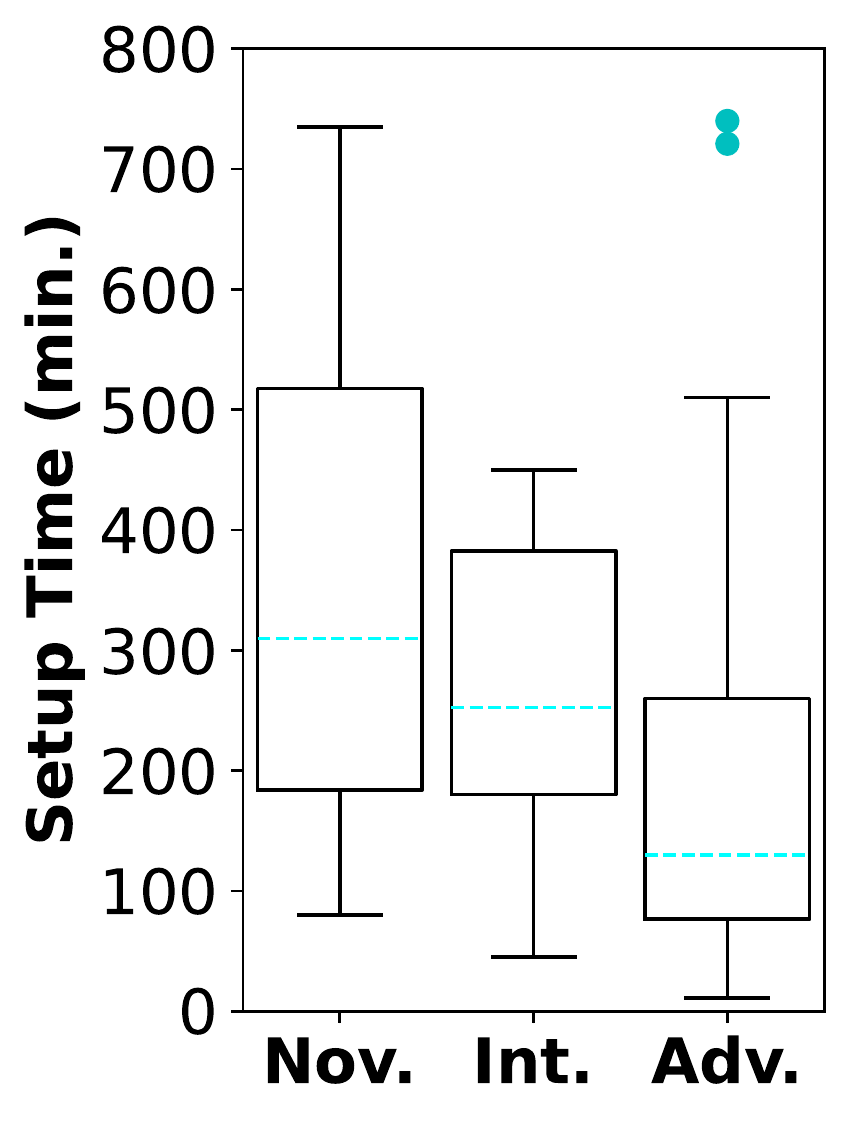}\hspace{-0.1em}\includegraphics[width=0.24\textwidth]{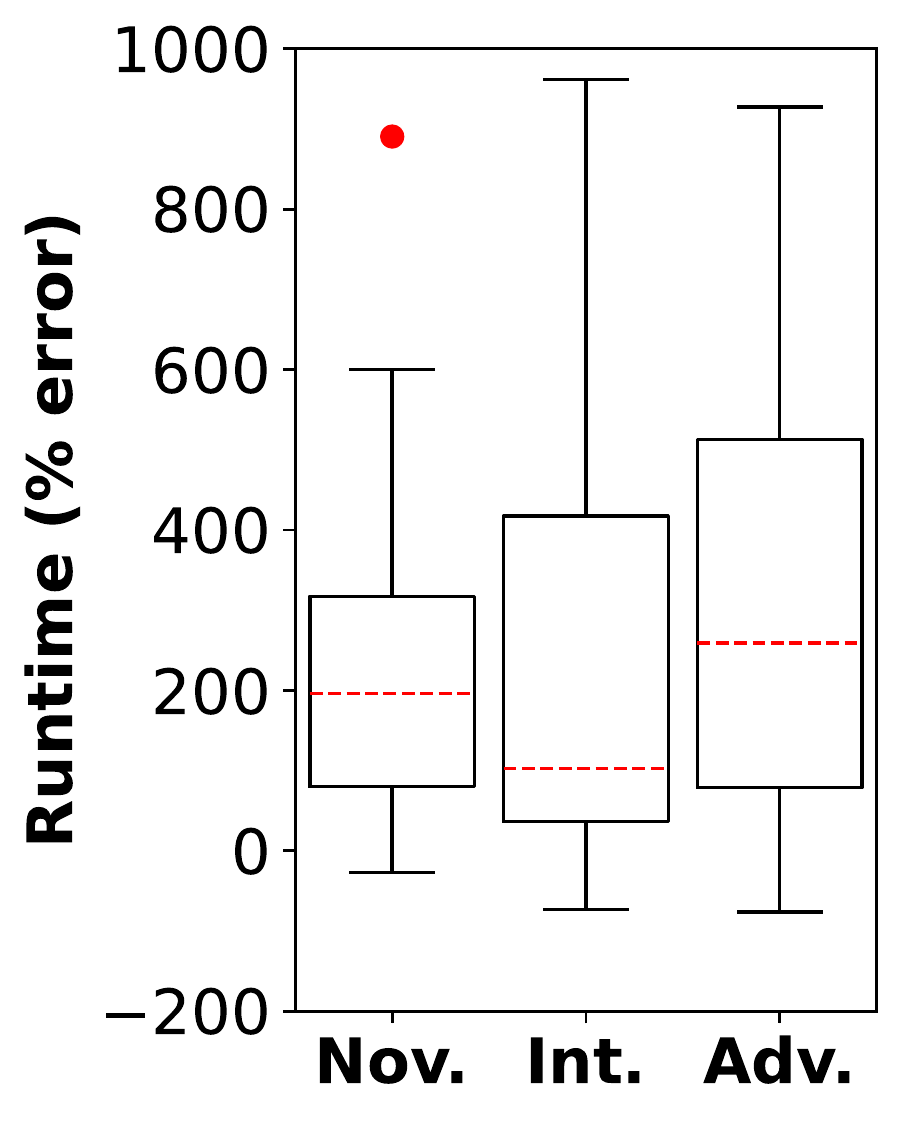}
    \vspace{-5pt}
    \caption{System setup time (minutes) and runtime (percent difference from research team's runtime) by skill level (novice, intermediate, or advanced).}
    \label{fig: setup time}
\end{figure}

\paragraph{Relationship with time.}
Figure~\ref{fig: setup time} shows the distribution of setup times and runtimes reported by students. We observe a striking variation in setup time across all skill levels, from under an hour to nearly 30 hours. 
As skill level increases, we observe that the median and minimum setup time, as well as the overall range of setup times, marginally decrease.
To examine how factors used to assign their skill level contribute to setup time, we calculate the Spearman correlation between each skill level factor and setup time in the second column of Table~\ref{tab: skill level vs code setup}.
Indeed, we observe a significant correlation, with understanding of the homework assignment on LSTMs having the strongest negative association with setup time. As this assignment required implementing and training language models in PyTorch, this suggests that these hands-on NLP skills may save students' time when reproducing NLP results. However, if we interpret $\rho^2$ as a coefficient of determination, {skill level factors} explain only up to $\rho^2=18.5\%$ of variance in setup time.\footnote{All variables have a high variance inflation factor, thus it is likely that they similarly contribute to a student's experience.} The large overlap in the setup time distribution between skill levels further suggests that there are more factors at play here.
Meanwhile, we see no clear differences in runtimes based on skill level, as each paper should have a consistent required runtime to train and evaluate models.

\begin{figure}
    \centering
    \includegraphics[width=0.48\textwidth]{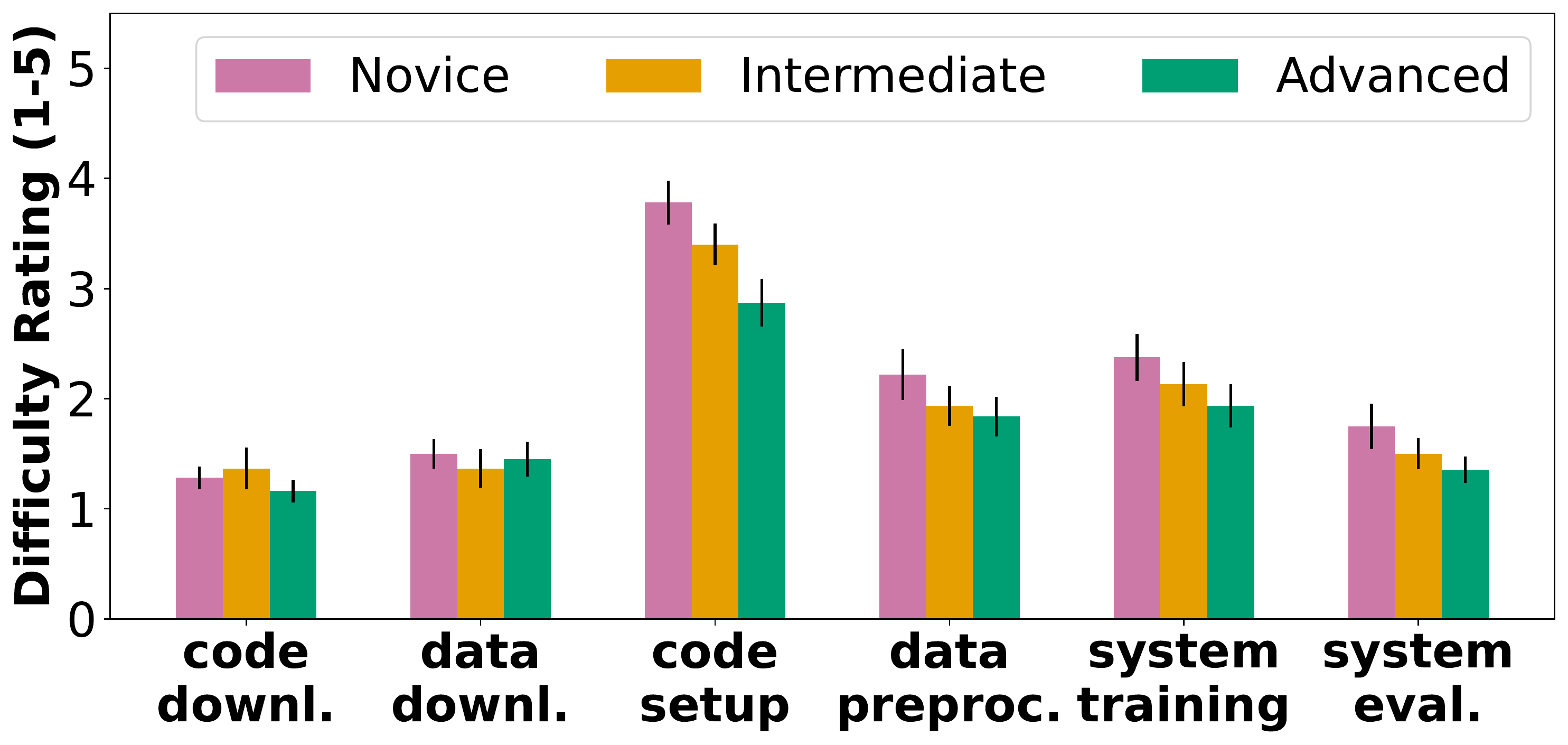}
    \caption{Mean reproducibility difficulty rating (1-5, 5 being most difficult) for each step of experiments: {downloading the code and data, setting up the code, preprocessing data, and training and evaluating the system.}}
    \vspace{-10pt}
    \label{fig: ease of reproducibility by skill level}
\end{figure}

\paragraph{Relationship with difficulty.}
For a more direct measure of students' experience, Figure~\ref{fig: ease of reproducibility by skill level} summarizes student ratings for the difficulty of each step of the experiment. For most steps of the experiment, more novice students reported slightly more difficulty. Students found code setup, data preprocessing, and system training to be the most difficult steps, and we observed a significant decrease in difficulty with respect to skill level for code setup.
This suggests a relationship between students' skill level and their reported code setup difficulty.

To understand this relationship, we calculate the Spearman correlation between each skill level factor and code setup difficulty rating, shown in the third column of Table~\ref{tab: skill level vs code setup}. Again, we find that all skill level factors are significantly correlated, with LSTM understanding again having the strongest association with lower reported difficulty.
Similarly, though, we observe a maximum $\rho^2 = 15.7\%$, suggesting that while some of students' reported difficulties may be explained by their skills, there is likely more to the story.
Further, it remains to be seen what exactly makes novice students feel worse about the difficulty of experiments, or why the rating for code setup is leaning negative overall. The remainder of our analysis provides possible answers to these questions.

\subsection{Student Comprehension of Paper}
Knowledge in NLP is primarily transferred through research papers. A student's ability to absorb knowledge from their assigned paper may relate to their ability to reproduce its results.
Here, we examine the relationship between their accuracies on paper comprehension questions in the post-survey and their experience, characterized by code setup time and difficulty rating, which exhibit the most significant variations across students.

\begin{figure}
    \centering
    \vspace{-8pt}
    \includegraphics[trim={1.3cm 0 0 0},clip,width=0.52\textwidth]{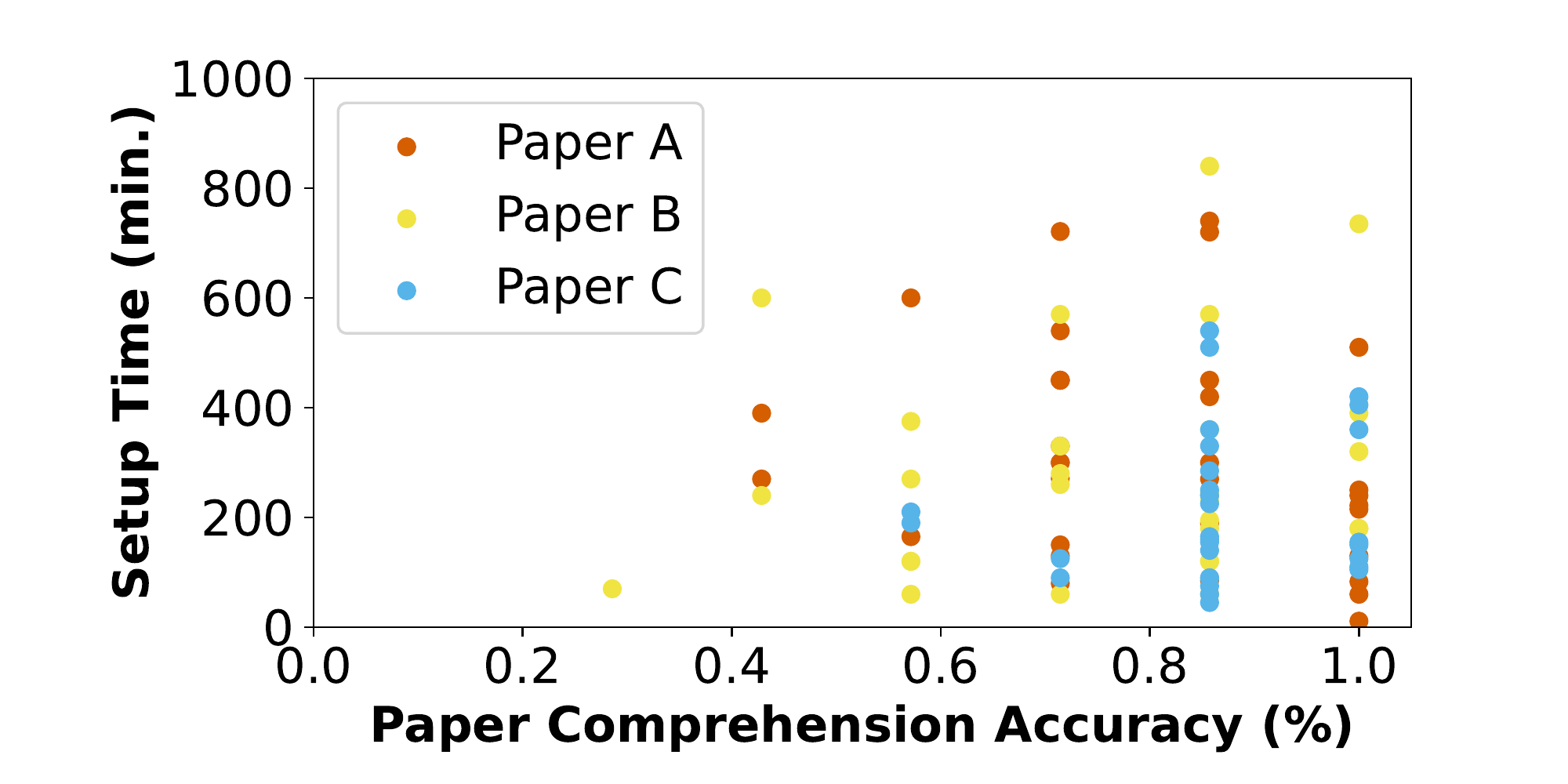}
    \vspace{-20pt}
    \caption{Students' accuracy on comprehension questions versus time to set up the experiment.}
    \label{fig: comprehension accuracy}
\end{figure}

As shown in Figure~\ref{fig: comprehension accuracy}, we observed a wide range of accuracies on these questions, with no clear correlation to their reported setup time. 
There is not a significant Spearman correlation between question accuracy and setup time or difficulty rating, suggesting that a student's comprehension of the work is not associated with their experience in reproducing its results.
This shows that even the clearest, most well understood paper may be difficult for beginners to engage with hands-on, and thus effective open-sourcing of code remains a separate and important issue to enable reproducibility.

\subsection{Author Reproducibility Efforts}\label{sec:bypaper}
Our results so far are vexing, as they show that students encounter vastly different levels of roadblocks in reproducing results that are only somewhat attributed to their skills, and unrelated to their comprehension of the work. To enable a frustration-free experience for all beginners, it is essential to understand what causes this disparity.
To investigate whether their experience instead depends on authors' specific efforts to make paper results reproducible, we examine the variations between the assigned papers in terms of time spent to reproduce results, as well as students' feedback on the experiments.

\begin{figure}
    \centering
    \hspace{-0.5em}\includegraphics[width=0.23\textwidth]{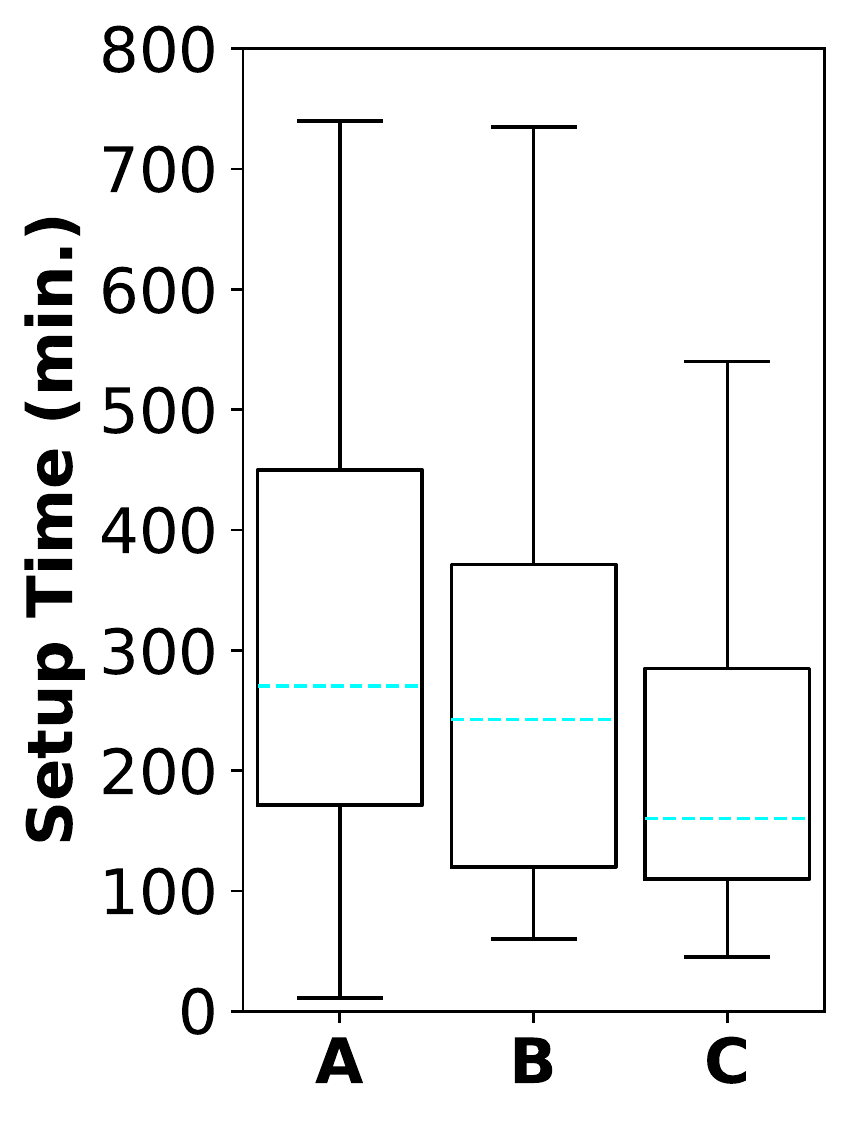}\hspace{-0.1em}\includegraphics[width=0.24\textwidth]{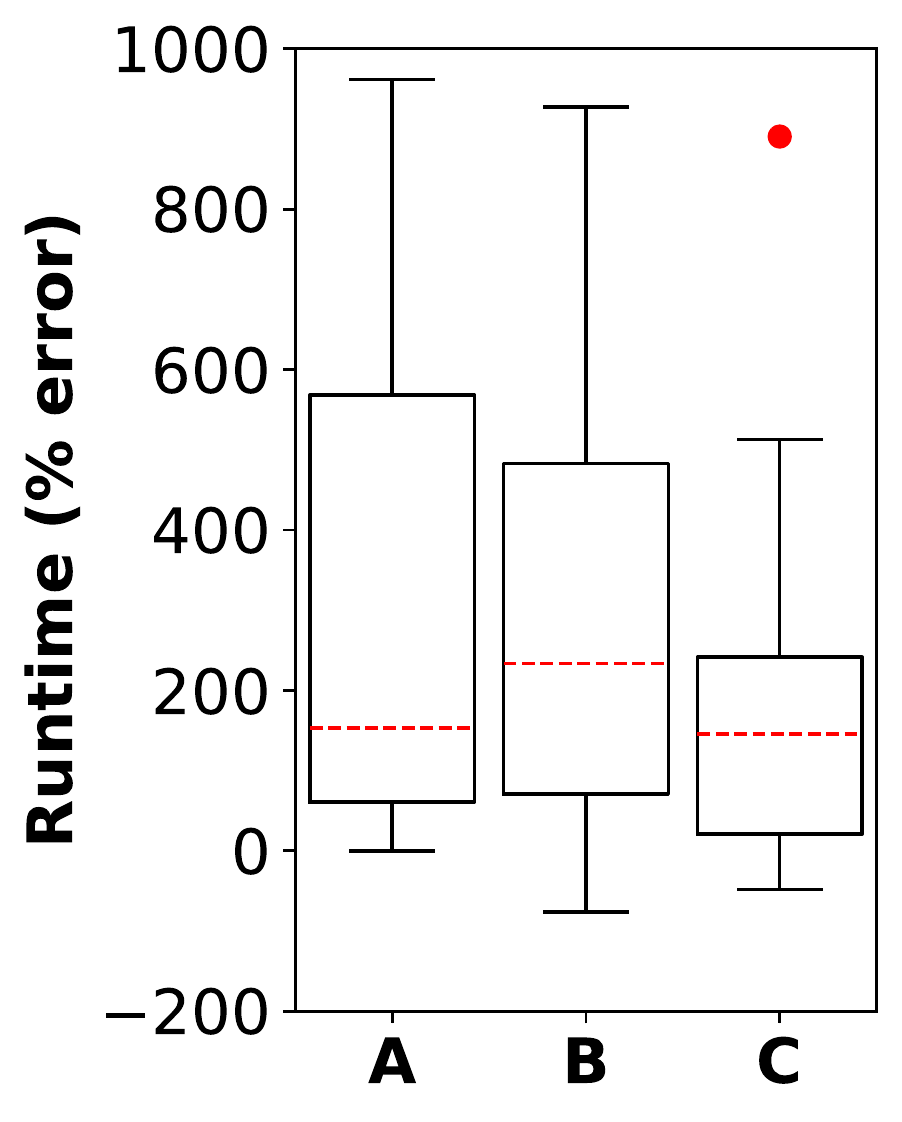}
    \vspace{-10pt}
    \caption{System setup time (minutes) and runtime (percent difference from research team's runtime) by assigned paper.}
    \vspace{-10pt}
    \label{fig: setup time by paper}
\end{figure}

\paragraph{Relationship with time.}
First, we examine the distribution of setup time and runtime by assigned paper in Figure~\ref{fig: setup time by paper}. A wide range of times is again observed for each paper, and the median setup time is consistently higher than the research staff's setup time of 2 hours, suggesting that compared to the experts on the research team, beginners are still learning to efficiently set up the code for an NLP experiment.
Meanwhile, the median runtime for all papers is also higher than than those of the research team, and a wide range of runtimes are again observed. This may suggest that across the board, {beginners encountered issues that caused them to troubleshoot by running code repeatedly}.
{Paper C's median and range of setup times, as well as its range of runtimes, were the smallest, possibly suggesting that the authors' efforts on reproducibility were especially helpful in enabling beginners to quickly reproduce their results.}

To understand these observed differences across papers, we take a closer look at reproducibility efforts for each paper.
After performing the experiment, students indicated in the post-survey which items from the ACLRC were most helpful in reproducing the results from their specific paper.\footnote{Items 2 and 17, which refer to the prerequisite steps of authors simply releasing their code and data, are omitted from analysis due to high variance inflation factor.} We analyze their responses with respect to setup time and runtime in Table~\ref{tab: checklist regression} by performing a multiple linear regression over items from the ACL Reproducibility Checklist as predictors for students' code setup time and runtime.

We find several significant correlations between checklist items and students' setup time and runtime. Specifically, reporting of best model hyperparameters (Paper A), model description (Paper B), and dataset partition information (Paper C) are all positively correlated with setup time, with hyperparameter bounds also positively correlated with runtime for Paper A. This may suggest that students encountered issues related to these factors that contributed to longer observed setup times and runtimes. Meanwhile, reporting of model selection information was associated with faster runtimes for Paper B, suggesting this was well addressed and important for a successful reproduction.


This analysis provides detailed insights about how each paper differed in terms of serving the needs of beginners.
Notably, the top-ranked reproducibility efforts can explain a relatively large amount of variance in setup time and runtime for some papers. {For example, $R^2=62\%$ of variance in Paper C's setup time and $66\%$ of variance in Paper B's runtime are explained by these efforts,} {providing some strong reasons for the wide ranges observed.}
This may suggest that these particular efforts are crucial for beginners' experience, and in general, that authors' reproducibility efforts can have a much larger impact on their experience than their technical skills or paper comprehension.

\begin{table}
    \centering
    \footnotesize

    \setlength\tabcolsep{5pt}
    \begin{tabular}{clrr}\toprule
        \textbf{Paper} & \textbf{Top ACLRC Item, Setup Time} & $\boldsymbol{\beta}$ &  $\boldsymbol{R^2}$ \\
         \cmidrule(lr){1-1} \cmidrule(lr){2-3} \cmidrule(lr){4-4}
            
         A & 10. Best Hyperparameters & \textbf{4.24} & 0.53 \\
         B & \hspace{5pt}1. Model Description & \textbf{8.47} & 0.15 \\
         C & 14. Dataset Partition Info & \textbf{4.08} & 0.62  \\\cmidrule(lr){1-1} \cmidrule(lr){2-3} \cmidrule(lr){4-4}
         All & \hspace{5pt}1. Model Description & 1.89 & 0.40 \\\bottomrule

    \end{tabular}

    \begin{tabular}{clrr}\toprule
        \textbf{Paper} & \textbf{Top ACLRC Item, Runtime} & $\boldsymbol{\beta}$ &  $\boldsymbol{R^2}$ \\
         \cmidrule(lr){1-1} \cmidrule(lr){2-3} \cmidrule(lr){4-4}
            
         A & \hspace{5pt}9. Hyperparameter Bounds & \textbf{46.43} & 0.17 \\
         B & 11. Model Selection Strategy & \textbf{-13.20} & 0.66 \\
         C & \hspace{5pt}6. Val. Set Metrics  & -3.26 & -0.04  \\\cmidrule(lr){1-1} \cmidrule(lr){2-3} \cmidrule(lr){4-4}
         All & \hspace{5pt}9. Hyperparameter Bounds & 6.61 & 0.07 \\\bottomrule

    \end{tabular}

    \normalsize
    \caption{Multiple linear regression over items from the ACL Reproducibility Checklist as predictors for students' code setup time and runtime (\% difference from research team). Most significant predictors for each assigned paper and overall are listed, along with their correlation coefficients $\beta$ and adjusted $R^2$ values. Statistically significant ($p<0.05$) coefficients in bold.}

    \label{tab: checklist regression}
\end{table}

\paragraph{Relationship with difficulty.}
In Table~\ref{tab: checklist regression difficulty}, we similarly analyze how papers' reproducibility efforts affected students' reported code setup difficulty, which earlier showed significant variations with respect to skill level.\footnote{See more about difficulty ratings for each paper in Appendix~\ref{apx: difficulty by paper}.} These results again indicate model hyperparameters (Paper A) and model selection strategy (Paper B) to be significantly associated with higher reported difficulty. Meanwhile, the reporting of model complexity was significantly associated with lower difficulty in Paper C, suggesting authors may have provided helpful information related to this. This may provide some more clues toward paper-specific variations in students' experience.

\begin{table}
    \centering
    \footnotesize
    \begin{tabular}{clr}\toprule
        \textbf{Paper} & \textbf{Top ACLRC Item, Setup Difficulty \:\:} & $\boldsymbol{\beta}$ \\
         \cmidrule(lr){1-1} \cmidrule(lr){2-2} \cmidrule(lr){3-3}
            
         A & 10. Best Hyperparameters & \textbf{1.82}  \\
         B & 11. Model Selection Strategy & \textbf{4.26}  \\
         C & \hspace{5pt}5. Model Complexity Info & \textbf{-4.40} \\ \cmidrule(lr){1-1} \cmidrule(lr){2-2} \cmidrule(lr){3-3}
         All & 15. Data Preprocessing Info & \textbf{0.65}  \\
        \bottomrule
        
    \end{tabular}
    
    \normalsize
    \caption{Ordinal logistic regression over items from the ACL Reproducibility Checklist as predictors for students' reported code setup difficulty (1-5, 5 being most difficult). Most significant predictors for each assigned paper and overall are listed, along with their correlation coefficients $\beta$. Statistically significant ($p<0.05$) coefficients in bold.}
    \vspace{-15pt}

    \label{tab: checklist regression difficulty}
\end{table}

\paragraph{Open-ended feedback.}
Beyond reproducibility efforts from the ACLRC, we surveyed students directly for additional open-ended feedback on their experience and outlooks on the experiments. 
We asked students what aspects of their assigned papers helped or blocked them in reproducing the results, as well as what should be added to the ACLRC to improve their experience.
We categorized student responses and aggregated them in Tables~\ref{fig: repro helpers}, \ref{fig: repro difficulties}, and \ref{fig: aclrc additions}. Comments focused on several aspects, varying by their assigned paper and unique experience with reproducing its results. Nonetheless, common responses provide rich insights about what NLP beginners need most to get their hands on the latest research. These insights, {primarily relating to engineering issues in using released code and data}, are summarized in Section~\ref{sec:discussion}. 

\begin{table}
    \centering
    \footnotesize
    \begin{tabular}{lcccc}\toprule
        \textbf{Reproducibility Helper} & \textbf{Frequency}\\
        \cmidrule(lr){1-2}
        {Clear Code Usage Documentation} & 56 \\
        {Example Scripts and Commands} & 27 \\
        {Easy-to-Read Code} & 15 \\
        {Easy-to-Access External Resources} & 13   \\
        {Sufficient Code Dependency Specification} & 12 \\
        \cmidrule(lr){1-2}
        {Other} & 11 \\
        \bottomrule
        
    \end{tabular}
    \normalsize
    \caption{Top 5 reported categories of features that helped students' reproduction of results. Less frequent responses aggregated in Other category.}
    \label{fig: repro helpers}
\end{table}


\begin{table}
    \centering
    \footnotesize
    \setlength{\tabcolsep}{4pt}
    \begin{tabular}{lcccc}\toprule
        \textbf{Reproducibility Blocker} & \textbf{Frequency}\\
        \cmidrule(lr){1-2}
        Insufficient Code Dependency Specification & 38 \\
        Difficult-to-Access External Resources & 27 \\
        Unclear Code Usage Documentation & 17 \\
        Pre-Existing Bugs in Code & 16 \\
        Difficult-to-Read Code & 11 \\
        \cmidrule(lr){1-2}
        {Other} & 30 \\
        \bottomrule
        
    \end{tabular}
    \normalsize
    \caption{Top 5 reported categories of features that blocked students' reproduction of results. Less frequent responses aggregated in Other category.}
    
    \label{fig: repro difficulties}
\end{table}


\begin{table}
    \centering
    \footnotesize
    \begin{tabular}{lcccc}\toprule
        \textbf{Suggested ACLRC Addition} & \textbf{Frequency}\\
        \cmidrule(lr){1-2}
        Standards for Documentation Clarity & 22 \\
        Full Specification of Code Dependencies & 18 \\
        Demonstration of Code Usage & 9 \\
        Provision of Support for Issues & 8 \\
        Standards for Code Clarity & 5 \\        
        \cmidrule(lr){1-2}
        {Other} & 23 \\
        Already Included & 23 \\
        \bottomrule
        
    \end{tabular}
    \normalsize
    \caption{Top 5 suggested categories of additions to the ACL Reproducibility Checklist (ACLRC). Less frequent suggestions and those already addressed in the ACLRC aggregated in Other and Already Included categories.}
    \label{fig: aclrc additions}
\end{table}


\section{Discussion \& Recommendations}\label{sec:discussion}
Our study reveals a wealth of insights into enhancing the accessibility of NLP research to beginners. The most interesting insight is that deliberate reproducibility efforts by authors beyond simply writing a paper and releasing the code are more crucial to beginners' experience in reproducing results than 
their programming skills and paper comprehension.
This finding behooves us researchers to diligently make these efforts, which would result in a win-win situation: people outside of the NLP research community, e.g., researchers from other disciplines and even the general public, can engage with our research more, which will extend the impact of our research.

Lastly, we share concrete, actionable recommendations on how to do this, framed around students' common feedback in Tables~\ref{fig: repro helpers}, \ref{fig: repro difficulties}, and \ref{fig: aclrc additions}. {Where we find that current reproducibility guidelines for NLP research are insufficient, we make recommendations on how they may be strengthened to consider beginners' experiences.}

\paragraph{Code dependencies.}
The most common complaint from students (reported for all papers) was the specification of code dependencies, e.g., the versions of Python and Python packages.
On the ACLRC, this effort is only briefly mentioned in Item 2, a general item about open-sourcing the code which does not reference version numbers of dependencies. Consequently, including more details about dependencies, especially version numbers, was the second most common suggestion by students to add to the ACLRC.
In contrast, the Responsible NLP Research checklist (RNLPRC) recently adopted at more ACL venues\footnote{\url{https://aclrollingreview.org/responsibleNLPresearch/}} emphasizes these efforts.
Fortunately, NLP researchers can rely on various computing environment management tools, such as \texttt{pip}\footnote{\url{https://pypi.org/project/pip/}}, \texttt{conda}\footnote{\url{https://docs.conda.io/en/latest/}}, {Poetry},\footnote{\url{https://python-poetry.org/}} and Docker.\footnote{\url{https://www.docker.com/}} Simply utilizing such tools when sharing our work can make a meaningful difference for beginners.

\paragraph{Instructions for reproducing results.}
Just releasing source code is not enough for others to reproduce results; it needs to be accompanied by clear usage documentation with steps to reproduce the results. {Documentation was the most appreciated effort by students, and also the third most common complaint, suggesting that it can make or break a beginner's experience in reproducing results. Standards for code documentation were the most common suggested addition to the ACLRC by students.} Good code documentation is a huge topic, and there are many resources available for this matter~\cite{Parnas2011,aghajani-2019-documentation,hermann2022documenting}.
Furthermore, students' specific feedback can also provide inspiration here. 
For example, one common suggestion for documentation standards on the ACLRC was clearly documenting the correspondence between code and results in the paper to highlight how to reproduce those results as reported.
Related to documentation, students' second most appreciated effort was providing examples for code usage, and the third most suggested addition to the ACLRC was a demonstration of code usage, whether it be an example script or command, an interactive notebook, or even a video. Neither the ACLRC or RNLPRC include specific recommendations for code usage instructions or documentation. {As such, we recommend that clear documentation of code usage be considered as a criterion, and it may be worthwhile for the ACL community to propose some standards for future work to follow.}

\paragraph{Availability of external resources.}
While all code was released on GitHub, large artifacts (e.g., datasets or pre-trained models) that are not suitable for \texttt{git} were hosted on other websites. Students reported difficulties when attempting to access them, such as broken links
and dependency on third-party software.
Conversely, students commonly rated ease of access to such resources helpful to reproducing results. While providing access to external resources is already suggested in the ACLRC, it is not explicitly mentioned in the RNLPRC, which may give an impression that this is not important, {despite being essential to reproducibility. As such, we recommend that this issue be discussed in the RNLPRC. Further, we suggest that NLP researchers should take extra care by using a centralized system like \texttt{HuggingFace Datasets}~\cite{lhoest-etal-2021-datasets},\footnote{\url{https://huggingface.co/docs/datasets}} or, at a minimum, periodically verifying that important resources are still accessible.}

\paragraph{Code clarity and functionality.}
Students found code clarity, including informative code comments and variable names, neatness of code, and intuitive file structures, to be the third most helpful effort in papers' code bases. While the ACLRC and RNLPRC make no recommendations for this, this was a common suggestion to add. Thankfully, there are widely agreed upon standards for code clarity~\cite{martin2009clean,wilson2014best}, and automated formatting tools like \texttt{black}\footnote{\url{https://black.readthedocs.io/en/stable/}} can make this easier.
Further, many students were blocked by minor bugs in the code when reproducing results. Authors should take extra care to avoid them in order to enable others to engage with their work without frustration. One common student suggestion for the ACLRC was to provide support for bugs, whether interactively through forums like GitHub Issues\footnote{\url{https://github.com/features/issues}} or proactively through an FAQ in their documentation. 
Another less common suggestion was to perform a sanity check reproduction of results on a clean copy of the code before open-sourcing and after substantial changes. Such an effort to mitigate even minor code bugs could make a substantial difference in the reproduction experience for beginners.

\vspace{10pt}

Addressing these key issues in NLP research practice\footnote{While our study focused on NLP research, many of the resulting insights are also applicable to neighboring disciplines.} could greatly improve the experience of beginners when reproducing experimental results, and extend the accessibility of the latest research developments to those even beyond the research community. {As NLP and AI research have recently attracted unprecedented global attention, we encourage the community to continue to dive deeper into the outlooks of beginners in future work. For example, given the recent paradigm shift in NLP from fine-tuning pre-trained language models to applying them directly to downstream tasks, there may also be a shift in user reproducibility roadblocks that will be important for the community to understand as it continues to strive for reproducible and accessible research. While some issues will remain important (e.g., code dependencies or data availability), other issues may become less crucial (e.g., hyperparameter search), while completely new issues may appear (e.g., choice of prompts used with a model). More broadly, there are also myriad opportunities to explore other topic areas and subject populations, the differences in reproducibility experiences between experts and beginners, and beginners' perception of state-of-the-art NLP systems and how they interact with them.}

\section*{Acknowledgements}
This research was supported in part through computational resources and services provided by Advanced Research Computing (ARC),\footnote{\url{https://arc.umich.edu/}} a division of Information and Technology Services (ITS) at the University of Michigan, Ann Arbor. We would like to thank the anonymous reviewers for their valuable comments and suggestions. We would also like to thank the authors of the three papers used in our study. Their exemplary work in sharing reproducible NLP experiments made this study possible.

\section*{Limitations}
\paragraph{Study scope.}
While our study only considers three papers, this is by design.
As our goal is to study user experience, by fixing papers to be within a specific topic area and time requirement, and having people with different skill levels reproduce the same papers, this allows us to have sufficient samples to understand general behaviors. It also blocks other nuance factors (e.g., introduced by different papers) on beginners' experience.
Each of our selected papers presented students with unique reproducibility barriers, and consequently resulted in a wealth of helpful insights. 
Furthermore, finding reproducible NLP papers that satisfied our constraints (as laid out in Section~\ref{sec: selection}) was surprisingly difficult, with only 3 out of 24 considered papers found to be reproducible within our constraints. 
Nevertheless, this study is still on the small scale. Engaging a larger community in a large scale study may provide additional insight.
Related to this, our study only includes a population of mostly graduate students at our university. Considering beginners from different educational backgrounds or regions could reveal more comprehensive insights, and we greatly encourage future efforts at a community level toward better understanding the needs of NLP beginners.

\paragraph{GPU runtime calculation.}
{It is also worth noting that it is difficult to consistently calculate runtime (as introduced in Section~\ref{sec:setup time runtime}) of code on GPU hardware, as fluctuations may occur due to a number of factors, including the specific GPU hardware allocated to a student,\footnote{{While experts used NVIDIA Tesla V100 GPUs with up to 16GB memory to reproduce results, NVIDIA A40 GPUs with up to 48GB memory are also available within the cluster.}} driver versions, and file systems experiments were run with. To minimize the impact of such issues, we chose to reproduce experiments that used small models and had shorter expected runtimes. Given that we observed runtimes up to several times larger than expert runtimes, we thus expect that trial and error in setting up experiments accounted for most fluctuation in observed runtimes.}

\section*{Ethics Statement}
This work was based on a study conducted as part of a graduate-level course including survey responses and computational resource usage statistics. 
Our institution’s Institutional Review Board (IRB) approved this human subjects research before the start of the study.\footnote{{Human subjects research approved by University of Michigan Health Sciences and Behavioral Sciences Institutional Review Board (IRB-HSBS), eResearch ID \texttt{HUM00224383}.}} {Subjects completed an informed FERPA-compliant consent form to opt into the study and were not compensated, since the collected data was part of a regular homework assignment.} As the research team for this work was also the instructional team of the course, one key ethical issue we aimed to mitigate was subjects feeling pressured to consent to this research in hopes it may benefit their grades. As such, we designated one member of the research team who was unable to view or modify student grades. Only this team member had access to informed consent responses from the students, and then linked and de-identified data before sharing it with the rest of the team. De-identification of data included classifying all free-text responses into a number of class labels so that students could not be recognized from their responses. Students were made aware that their participation was entirely optional, and could not possibly impact their grade due to this careful arrangement.
Further, to ensure that students were assigned a comparable amount of work, we carefully selected papers with results that could be reproduced by the research staff in a comparable amount of time (i.e., 2 hours).\footnote{As mentioned in Appendix~\ref{apx: paper selection}, one considered paper was discarded due to taking a significantly shorter amount of time than the others.}

The results of this study could have a positive impact on the NLP research community, as it reveals insights that may be helpful for NLP researchers to better enable beginners to get their hands on research artifacts and reproduce their results. If applied in future open-sourcing of research artifacts, such insights could expand the accessibility of our work to a broader audience. As NLP systems are becoming more ubiquitous in society and attracting attention beyond our research community, this effort could result in the inclusion of more voices in discussions around them and their future development, which is essential for democratization.

\bibliography{main}
\bibliographystyle{acl_natbib}

\clearpage

\appendix

\section{ACL Reproducibility Checklist}\label{apx: acl checklist text}


The full ACL Reproducibility Checklist is provided below.

\begin{itemize}[leftmargin=*]
\item For all reported experimental results:
    \begin{enumerate}[leftmargin=*]
    \setcounter{enumi}{0}
    \item A clear description of the mathematical setting, algorithm, and/or model;
    \item A link to (anonymized, for submission) downloadable source code, with     specification of all dependencies, including external libraries;
    \item A description of the computing infrastructure used;
    \item The average runtime for each model or algorithm, or estimated energy cost;
    \item The number of parameters in each model;
    \item Corresponding validation performance for each reported test result;
    \item A clear definition of the specific evaluation measure or statistics used to
    report results.
    \end{enumerate}
\item For all results involving multiple experiments:
    \begin{enumerate}[leftmargin=*]
    \setcounter{enumi}{7}
        \item The exact number of training and evaluation runs;
        \item The bounds for each hyperparameter;
        \item The hyperparameter configurations for best-performing models;
        \item The method of choosing hyperparameter values (e.g., manual tuning, uniform sampling, etc.) and the criterion used to select among them (e.g., accuracy);
        \item Summary statistics of the results (e.g., mean, variance, error bars, etc.).
    \end{enumerate}
\item For all datasets used:
    \begin{enumerate}[leftmargin=*]
    \setcounter{enumi}{12}
        \item Relevant statistics such as number of examples and label distributions;
        \item Details of train/validation/test splits;
        \item An explanation of any data that were excluded, and all pre-processing steps;
        \item For natural language data, the name of the language(s);
        \item A link to a downloadable version of the dataset or simulation environment;
        \item For new data collected, a complete description of the data collection process, such as ownership/licensing, informed consent, instructions to annotators and methods for quality control.
    \end{enumerate}
\end{itemize}

\section{Supplementary Results}\label{sec: supp results}

Here, we include some extra results that, while significant or informative, were less relevant to the message of our paper.

\subsection{Reproduced Accuracy}\label{apx: relative error}
\begin{figure}
    \centering
    \includegraphics[width=0.47\textwidth]{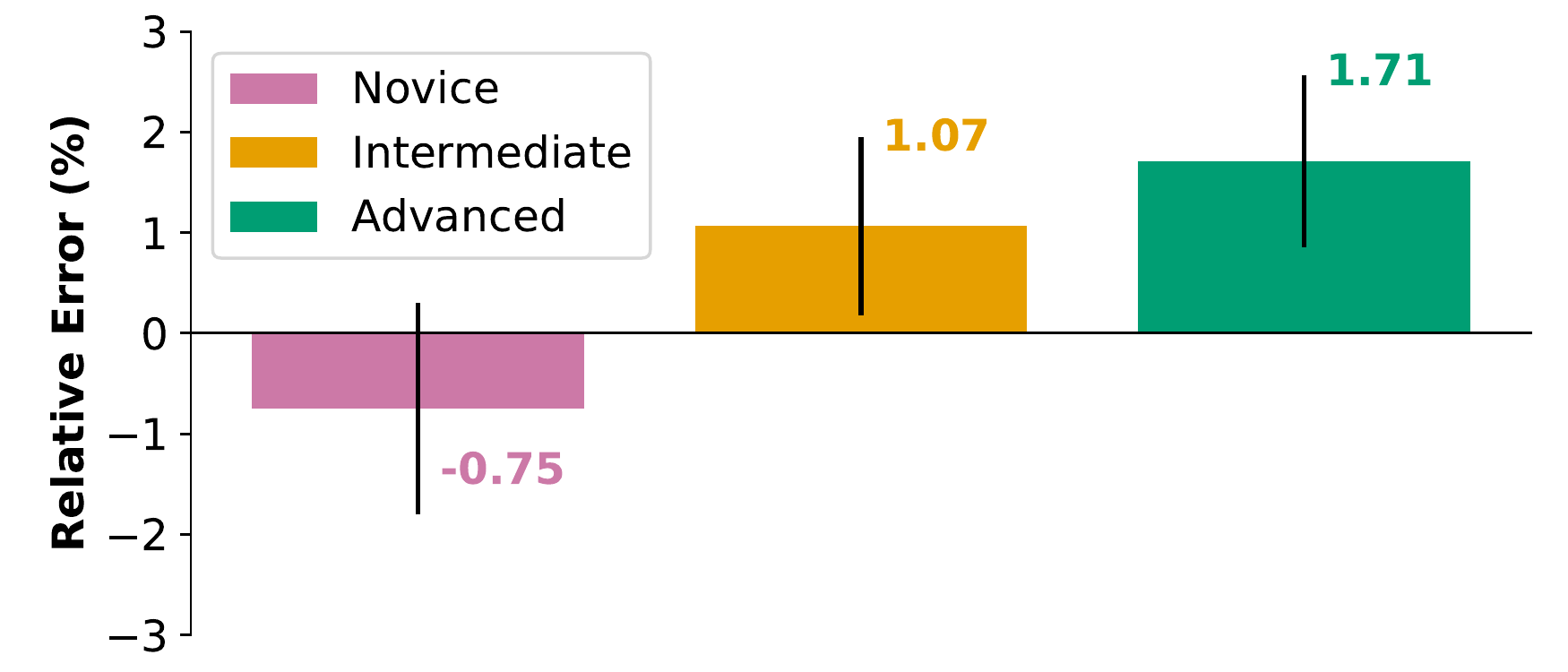}
    \vspace{-5pt}
    \caption{Average reported relative error from paper results, by students' technical skill level.}
    \vspace{-5pt}
    \label{fig: relative error}
\end{figure}

For each paper, we asked students to re-train one NLP system and report the accuracy in the same settings and on the same data partitions as reported in the paper. Figure~\ref{fig: relative error} averages the relative error of students' submitted results by experience group. As shown, on average, student results do not vary significantly by experience level. Further, on average, student results come close to those reported in the paper, and for the most part, do not differ significantly from our reproduced results.

\begin{figure}
    \centering
    \includegraphics[width=0.47\textwidth]{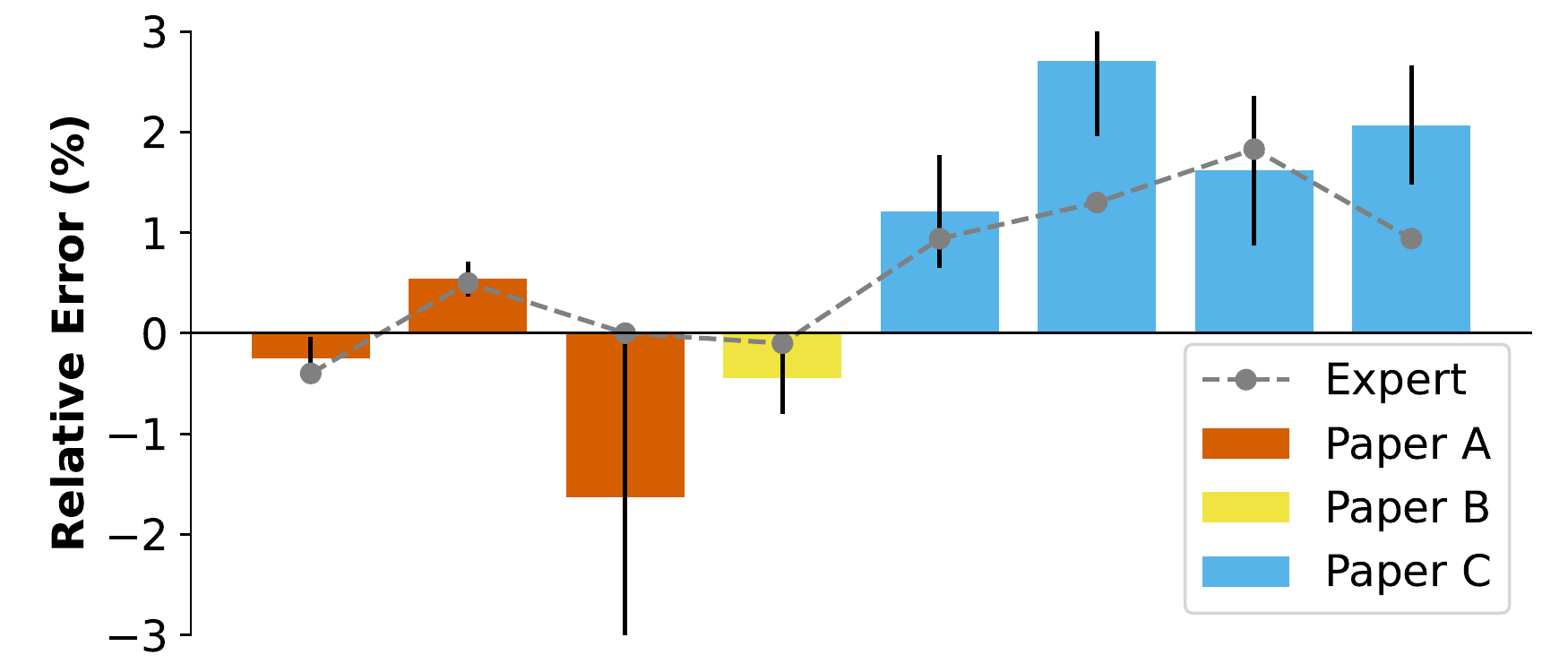}
    \caption{Relative error of students' reported results from those reported in the paper for each evaluation setting, compared to our expert result (dotted line).}
    \label{fig: accuracy_paper}
\end{figure}

In Figures~\ref{fig: relative error} and \ref{fig: accuracy_paper}, we compare the relative error between students' reported results for each skill level, and for each setting of the reproduced system and the results published in their corresponding papers.
In both cases, the student-obtained results are fairly aligned to the reported results in the paper (as well as our own reproduced results), with standard errors within 3\% accuracy.

\subsection{Reproducibility Difficulty by Paper}\label{apx: difficulty by paper}
\begin{figure}[H]
    \centering
    \hspace{-3pt}\includegraphics[width=0.48\textwidth]{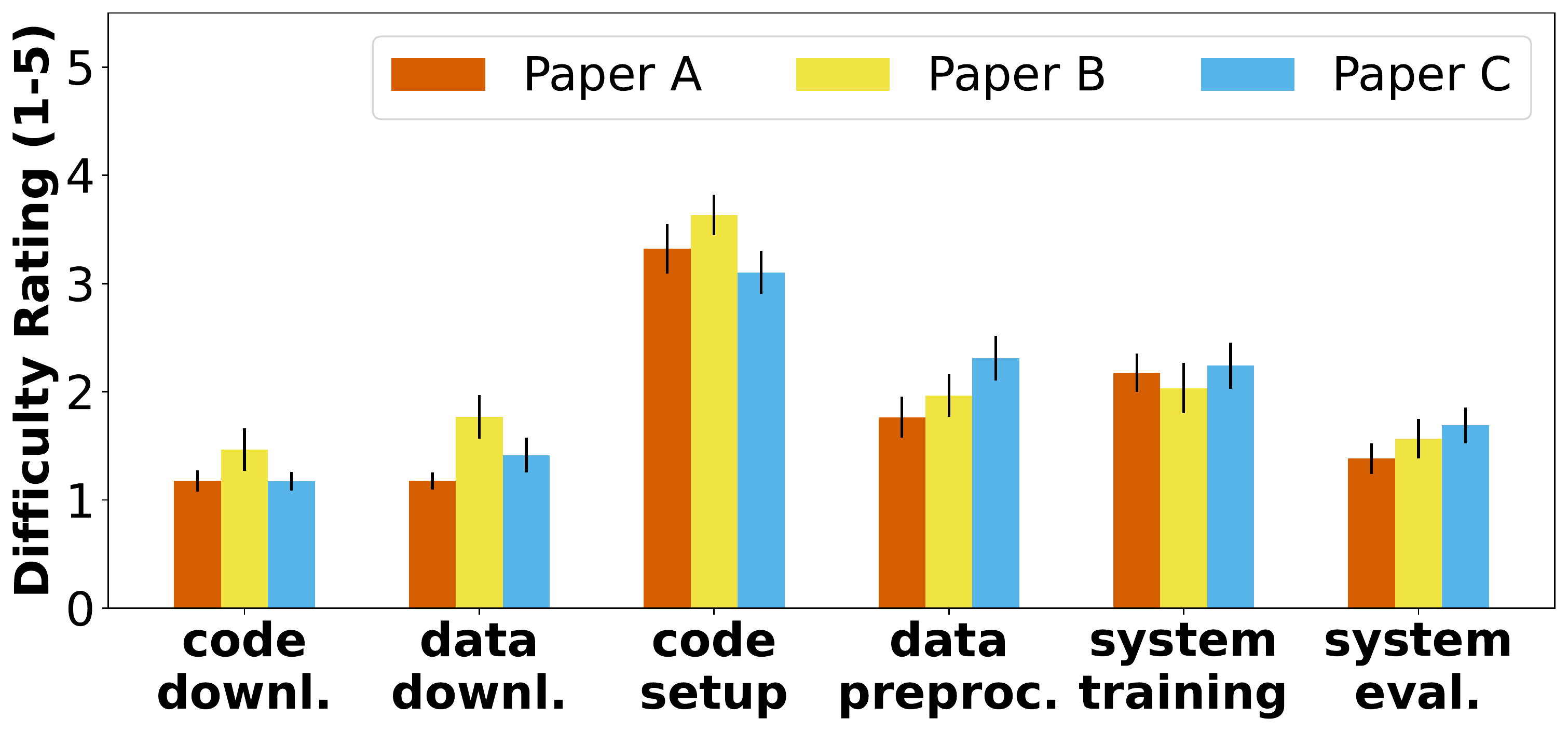}
    \caption{Mean reproducibility difficulty rating (1-5, higher being most difficult) for each step of experiments.}
    \vspace{-10pt}
    \label{fig: ease of reproducibility}
\end{figure}

Figure~\ref{fig: ease of reproducibility} summarizes student ratings for how difficult each step of the exercise was. 
Code setup, data preprocessing, and system training unsurprisingly had the lowest ratings, and mean ratings did not vary much across papers.

\subsection{ACL Reproducibility Checklist Survey}\label{apx: checklist items}
\begin{figure*}
    \centering
    \hspace{-0.5em}\includegraphics[width=0.99\textwidth]{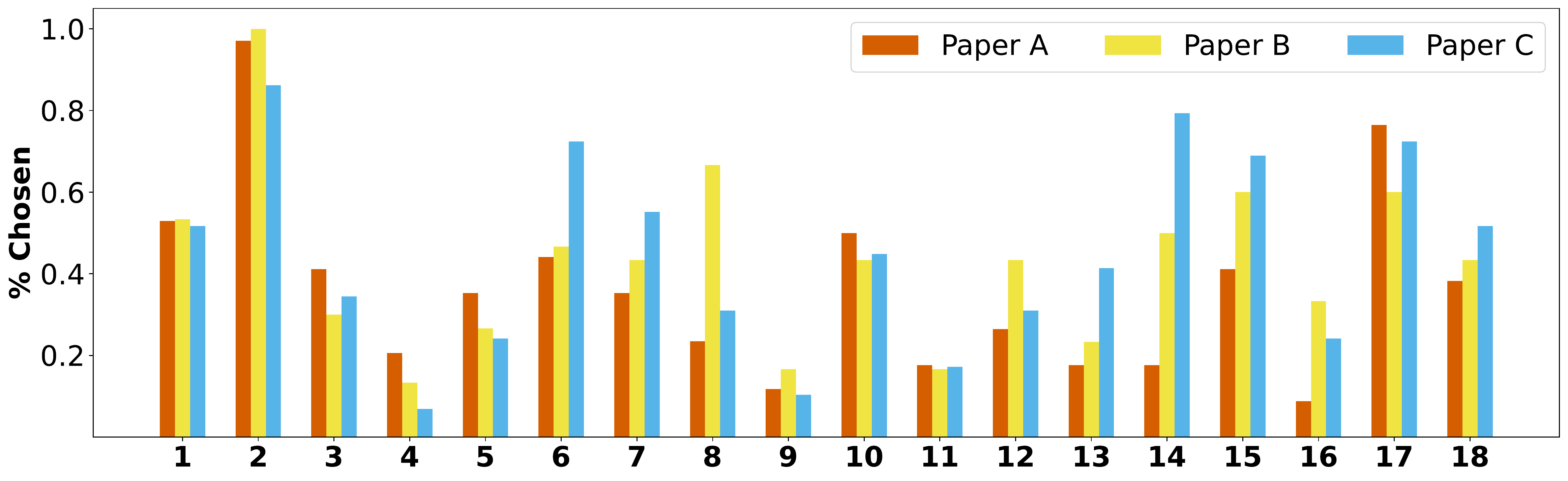}
    \caption{Percentage of responses in which each ACL Reproducibility Checklist item (indexed by the ordering of items) was selected as helpful by students in the study, by paper. See Appendix~\ref{apx: acl checklist text} for full text of checklist.}
    \label{fig:repro checklist barchart}
\end{figure*}

After completing their assigned work, students indicated which items on the ACL Reproducibility Checklist were most helpful in reproducing the results. In Figure~\ref{fig:repro checklist barchart}, we show the percentage of times each item was selected, aggregated by paper. The differences between items in the graph may give further insights on which parts of the checklist are most helpful to NLP beginners, and where our studied papers differed in terms of what was provided in open-sourced materials.

\section{Paper Selection}\label{apx: paper selection}
To find experiments to reproduce for the study, we considered papers from ACL conferences, which are top venues for NLP research. We specifically collected papers from the Annual Meeting of the ACL, the Conference on Empirical Methods in Natural Language Processing (EMNLP), and the Conference of the North American Chapter of the Association for Computational Linguistics: Human Language Technologies (NAACL HLT) from the years 2018 through 2022. Over these conferences, we arbitrarily selected 24 long papers from two topic areas: (A) \textit{Semantics: Sentence-level Semantics and Textual Inference},\footnote{Prior to EMNLP 2020, this area was referred to as \textit{Semantics: Textual Inference and Other Areas of Semantics}, but was merged with another area.} and (B) \textit{Language Generation}. {All selected papers had publicly available code, models (where applicable), and data with licenses permitting academic research use.}

For each selected paper, we attempted to reproduce the results of an experiment ourselves. If a paper presented several experiments, e.g., for different model variations, we chose the best-performing instance based on the evaluation metrics applied. If the best-performing instance could not be reproduced within a feasible amount of time or compute resources, we chose the next best that fit within our computing restrictions.
For our expert reproduction of results, we limited the effort to 2 hours of setup time and 4 hours of total code runtime (for successful runs of the code) on a single GPU\footnote{Setup time and runtime defined in Section~\ref{sec:setup time runtime}.} for any model training or evaluation, while our computing restrictions were dictated by the available hardware and software within the environment used to reproduce results.\footnote{For reproducing all results, both experts and students used an onsite computing cluster that offers access to NVIDIA Tesla V100 GPUs with up to 16GB memory, {and NVIDIA A40 GPUs with up to 48GB memory}.}

Out of these 24 papers, the results of only 4 papers could be successfully reproduced within the above constraints, all of which belonged to Area A. As this research was conducted as part of a homework assignment, we discarded one of these 4 papers that took a significantly shorter time to reproduce than the others.
Common reasons that the experiment failed included long model training time requirements, incompatibility of code bases with our computing platform, incompleteness of documentation, and discrepancy between reproduced results and those reported in papers.

{The specific papers we selected for the study are listed in Table~\ref{tab: paper info}, and the distribution of paper assignments across skill levels is listed in Table~\ref{tab: paper assignments}. In Appendix~\ref{apx: hw content}, we describe the specific experiments that we reproduced, and thus assigned for students to reproduce.}

\section{Pre-Survey Questions}\label{apx: presurvey}

As discussed in Section~\ref{sec:skill level} we conducted a brief survey on students' background at the beginning of the study. As mentioned in Section~\ref{sec: skill level assignment}, the results of this survey were used in characterizing students' skill levels. Pre-survey questions relevant to the study are listed below.

\begin{enumerate}[leftmargin=*]
    \item Before taking this course, how many years of experience did you have specifically with Python?
    \item Before taking this course, how many years of experience did you have specifically with PyTorch?
    \item To the best of your memory, how difficult was it for you to complete these past homework problems?
    
    \textit{Included pointers to specific homework problems on implementing LSTMs and transformers, and asked about them individually.}
    
    \begin{itemize}[leftmargin=*]
        \item \textit{very difficult}
        \item \textit{slightly difficult}
        \item \textit{neutral}
        \item \textit{slightly easy}
        \item \textit{very easy}
    \end{itemize}
\end{enumerate}

\section{Homework Assignment Content}\label{apx: hw content}
{We include some content from the homework assignment associated with this study which was used to prepare students for the work and prime them to report their results in the post-survey. Among course-related administrative details (e.g., how to access surveys, grading guidelines, etc.), we gave them important information on which experiment to reproduce from their assigned papers, what steps they should take when reproducing the results, and how to track their time spent on the experiment.}

\paragraph{Assigned experiments.}
{For each paper, the specific experiments we reproduced (and instructed students to reproduce) were as follows.}

{For Paper A, we instructed students to reproduce the following experiment from \citet{zhou-etal-2021-temporal}: \textit{Fine-tune \textsc{PtnTime} model to the uniform-prior TRACIE dataset, then evaluate it on the testing set (\textsc{PtnTime} result with 76.6\% "All" accuracy in Table 2 from the paper). {Report the Start, End, and All accuracies on the testing set.}}}

{For Paper B, we instructed students to reproduce the following experiment from \citet{donatelli-etal-2021-aligning}: \textit{Train the base alignment model with \textsc{BERT} embeddings on the pre-tagged Action Alignment Corpus, and evaluate on the testing set ("Our Alignment Model (base)" result in Table 4). {Report the combined accuracy on the testing set after cross-validation.}}}

{For Paper C, we instructed students to reproduce the following experiment from \citet{gupta-etal-2020-infotabs}: \textit{Train \textsc{RoBERTa}-base on the InfoTabs dataset with \textsc{TabFact} structured premise representation, and evaluate it on the development set and all three testing sets (\textsc{RoBERTaB}/\textsc{TabFact} result with 68.06\% Dev accuracy in Table 7 from the paper). {Report the accuracy on the development set and all three testing sets.}}}

\paragraph{Reading the paper.} 
\textit{The first step for this assignment is to read the paper. This will make clear the experiments and
results the paper has contributed, so that when you begin to reproduce the results, you will already
know what to expect.}

\paragraph{Reproducing the results.} 
\textit{Next, you should open the code base for your paper and follow the authors’ instructions for re-
producing the results (only those experiments that we specify). Their documentation will usually
include some combination of the following steps:}
\begin{itemize}[leftmargin=*,noitemsep]
    \item \textit{Cloning a GitHub repo to your local environment (make sure you have an account)}
    \item \textit{Installing Python dependencies}
    \item \textit{Downloading datasets and pre-trained models}
    \item \textit{Training and validating models}
    \item \textit{Testing models}
\end{itemize}

\textit{Different papers will give varying degrees of guidance for these steps. Documentation for a paper’s code release may have typos or missing information. You should use your understanding of the
paper, Python, and PyTorch to fill in any gaps.}

\paragraph{Tracking your time.}
\textit{Please carefully track the time you spend on the homework assignment. Categorize your time into
the following activities:}
\begin{enumerate}[leftmargin=*,noitemsep]
    \item \textit{Closely \textbf{reading} the paper}
    \item \textit{Setting up the \textbf{code base} to successfully run experiments, e.g., downloading the code and data, preprocessing data, and installing dependencies}
    \item \textit{\textbf{Training} models (i.e., waiting for training code to run)}
    \item \textit{\textbf{Evaluating} models (i.e., waiting for evaluation code to run)}
    \item \textit{\textbf{Documenting} results and completing the post-survey    }
\end{enumerate}

\textit{You will be asked for this information in the post-survey to help us understand the
difficulties that you ran into. Please report your time honestly; your grade will not depend on how
much time it took to complete the assignment.}

\section{Post-Survey Questions}\label{apx: postsurvey}

As discussed in Section~\ref{sec:post-survey}, students were asked to complete a survey after reproducing their assigned results.\footnote{{Students were primed for some of these questions through assignment documents we distributed to introduce the expected work and students' assigned papers and experiments. Relevant content is included in Appendix~\ref{apx: hw content}.}} Post-survey questions relevant to the study are listed below.

\begin{enumerate}[leftmargin=*]
    \item \textit{A series of paper-specific reading comprehension questions, {listed in Appendix~\ref{apx: comprehension questions}}.}
    \item Were you able to reproduce the results of the paper?
    \item Please report the performance of your trained model under each of the evaluation settings. 

    \textit{See Appendix~\ref{apx: hw content} for specific results reported for each paper.}
    
    \item How much time did it take you to set up the code base for model training? Count from the time the code base was downloaded until model training was successfully running, including data preprocessing. Don't count time spent actually running training code. \textit{(in hours and minutes)}
    \item How much time did it take you to set up the code base for model evaluation? Count from the time the code base was downloaded until model evaluation was successfully running, including data preprocessing. Don't count time spent actually running evaluation code. \textit{(in hours and minutes)}
    \item Please rate the level of difficulty you encountered in each step of the experiment. \textit{(1: very difficult - 5: very easy)}
    \begin{itemize}[leftmargin=*]
        \item \textit{Downloading the code base.}
        \item \textit{Downloading required data.}
        \item \textit{Setting up the code base and its dependencies.}
        \item \textit{Preprocessing the data.}
        \item \textit{Training models.}
        \item \textit{Evaluating models.}
    \end{itemize}
    \item Did the authors provide anything in the code release that helped make the results easier for you to reproduce?
    \item What could the authors have provided or done better for you to reproduce the results faster or with less frustration?
    \item Which of the following items from the ACL Reproducibility Checklist were especially helpful for reproducing the paper's results? Please check all that apply.

    \textit{Checklist items listed in Appendix~\ref{apx: acl checklist text}.}

    \item Is there anything you would add to this checklist that you wish authors would provide to help you reproduce results faster or with less frustration? Suggest up to 5.
\end{enumerate}

\section{Comprehension Questions}\label{apx: comprehension questions}
{Here, we list the specific comprehension questions used in the post-survey for each paper (correct answers in bold).}

\subsection{Paper A Questions}
\begin{enumerate}[leftmargin=*]
    \item \textbf{Motivation:} Which of the following is not a motivation of this work?
    \normalsize   
    \begin{enumerate}[leftmargin=*]
        \item \textit{Humans can recognize temporal relationships between events both explicitly mentioned and implied (but not explicitly mentioned) in language.}
        \item \textit{Past work in temporal reasoning has focused only on explicitly mentioned events.}
        \item \textbf{\textit{At the time of writing, there were no benchmark datasets evaluating temporal reasoning for NLP.}}
        \item \textit{Constructing a latent timeline of events is essential for NLP systems to understand stories.}
    \end{enumerate}
    \normalsize

    \item \textbf{Problem Definition:} What task is being studied in this paper?
    \normalsize   
    \begin{enumerate}[leftmargin=*]
        \item \textbf{\textit{Textual entailment with a focus on ordering of events.}}
        \item \textit{Question answering with a focus on temporal reasoning.}
        \item \textit{Story generation with a coherent timeline of events.}
        \item \textit{Semantic role labeling for implicit events.}
    \end{enumerate}
    \normalsize

    \item \textbf{Approaches:} What are the inputs to the \textsc{PtnTime} system?
    \normalsize   
    \begin{enumerate}[leftmargin=*]
        \item \textit{A short story, a timeline of explicit and implicit events, and a proposed ordering of events.}
        \item \textit{A story written in text, a hypothesis written in text, and an inference label.}
        \item \textit{A story written in text, with BIO tags describing the spans of text where a specific event occurs.}
        \item \textbf{\textit{A story written in text, and a hypothesis about the temporal ordering of events in the story.}}
    \end{enumerate}
    \normalsize

    \item \textbf{Approaches:} What are the outputs of the \textsc{PtnTime} system?
    \normalsize   
    \begin{enumerate}[leftmargin=*]
        \item \textit{Tags describing the span of text the target event occurs.}
        \item \textbf{\textit{An inference label of entailment or contradiction.}}
        \item \textit{An inference label of entailment, neutral, or contradiction.}
        \item \textit{A timeline of explicit and implicit events in the text.}
    \end{enumerate}
    \normalsize

    \item \textbf{Implementation:} Which of the following files from the paper's code base is responsible for fine-tuning \textsc{PtnTime} to TRACIE?
    \normalsize   
    \begin{enumerate}[leftmargin=*]
        \item \texttt{tracie/code/models/symtime/ train\_t5.py}
        \item \texttt{tracie/code/models/ptntime/ evaluator.py}
        \item \textbf{\texttt{tracie/code/models/ptntime/ train\_t5.py}}
        \item \texttt{tracie/code/models/ptntime/ train\_ptntime.py}
    \end{enumerate}
    \normalsize
        
    \item \textbf{Results:} What is the meaning of the values in the "Story" column in Table 1 of the paper?
    \normalsize   
    \begin{enumerate}[leftmargin=*]
        \item \textit{The average percentage of hypotheses which were correctly classified for each story.}
        \item \textit{The percentage of stories which were correctly classified before a hypothesis was introduced.}
        \item \textit{The percentage of stories for which the system made a correct prediction on any hypothesis.}
        \item \textbf{\textit{The percentage of stories for which the system made correct predictions on all hypotheses.}}
    \end{enumerate}
    \normalsize
    
    \item \textbf{Conclusion:} Which of the following CANNOT be concluded based on the results in this paper?
    \normalsize   
    \begin{enumerate}[leftmargin=*]
        \item \textit{Symbolic temporal reasoning can be used to improve language models’ temporal reasoning.}
        \item \textit{Distant supervision on event durations can be used to improve language models' temporal reasoning.}
        \item \textit{The uniform-prior setting of TRACIE is harder for baseline systems to solve than the full dataset.}
        \item \textbf{\textit{When used in a zero-shot setting, the proposed models consistently outperform baselines from prior work fine-tuned on the task.}}
    \end{enumerate}
    \normalsize
        
    \end{enumerate}

\subsection{Paper B Questions}

\begin{enumerate}[leftmargin=*]
    \item \textbf{Motivation:} Which of the following is NOT a motivation of this work?
    \normalsize   
    \begin{enumerate}[leftmargin=*]
        \item \textit{A key challenge with interpreting cooking recipes is that for any dish, different recipes may omit or emphasize different steps.}
        \item \textbf{\textit{Aligning recipes based only on actions ignores rich information about the structure of recipes and relationships between sentences.}}
        \item \textit{Aligning multiple recipes for a single dish would give a recipe understanding system more complete information about making that dish.}
        \item \textit{Past work has not looked into aligning multiple recipes at the action level.}
    \end{enumerate}
    \normalsize
    
    \item \textbf{Problem Definition:} What task is being studied in this paper?
    \normalsize   
    \begin{enumerate}[leftmargin=*]
        \item \textbf{\textit{Alignment of actions in multiple recipes for the same dish.}}
        \item \textit{Alignment of sentences in multiple recipes for the same dish.}
        \item \textit{Alignment of actions in recipes for similar (but not identical) dishes.}
        \item \textit{Alignment of sentences across recipes for similar (but not identical) dishes.}
    \end{enumerate}
    \normalsize
    
    \item \textbf{Approaches:} What are the inputs to the base alignment model?
    \normalsize   
    \begin{enumerate}[leftmargin=*]
        \item \textit{One sentence describing one or more actions in a recipe.}
        \item \textit{A span of steps in a recipe, each of which is a sentence.}
        \item \textbf{\textit{Two recipes for a single dish, and a source action selected from one of the recipes.}}
        \item \textit{Two recipes for two different dishes, and a source action selected from one of the recipes.}
    \end{enumerate}
    \normalsize
    
    \item \textbf{Approaches:} What are the outputs of the base alignment model?
    \normalsize   
    \begin{enumerate}[leftmargin=*]
        \item \textit{All actions from the target recipe such that their confidence scores for aligning to the given source action exceed a threshold value.}
        \item \textit{A single action from the target recipe which best aligns to the given source action.}
        \item \textbf{\textit{Either one action from the target recipe which best aligns to the given source action, or no matching actions.}}
        \item \textit{The top five actions from the target recipe which best align to the given source action.}
    \end{enumerate}
    \normalsize
    
    \item \textbf{Implementation:} In which file are confidence scores from the alignment model calculated?
    \normalsize   
    \begin{enumerate}[leftmargin=*]
        \item \texttt{ara/Alignment\_Model/main.py}
        \item \texttt{ara/Alignment\_Model/utils.py}
        \item \texttt{ara/Alignment\_Model/ training\_testing.py}
        \item \textbf{\texttt{ara/Alignment\_Model/model.py}}
    \end{enumerate}
    \normalsize
    
    \item \textbf{Results:} How are the alignment model results from Table 4 in the paper calculated?
    \normalsize   
    \begin{enumerate}[leftmargin=*]
        \item \textit{The model is trained on the training set, validated on a validation set consisting of recipes for the same dishes as the training set, then tested on a set of recipes for dishes not seen in training or validation. The accuracy on this 
        testing set is reported in Table 4.}
        \item \textbf{\textit{Ten instances of the alignment model are trained using cross validation, where each fold holds out the recipes from one of the ten dishes as validation data, and from another dish as testing data. The testing results on each of the ten dishes are combined from the ten model instances.}}
        \item \textit{The model is trained on the training set, validated on a validation set consisting of recipes for the same dishes as the training set, then tested on a set of held-out recipes for those same dishes. The accuracy on this testing set is reported in Table 4.}
        \item \textit{Ten instances of the alignment model are trained using cross validation, where each fold holds out one of the ten dishes as testing data, and a validation set of recipes is randomly sampled from the training data. The testing results on each of the ten dishes are combined from the ten model instances.}
    \end{enumerate}
    \normalsize

    \item \textbf{Conclusion:} Which of the following CANNOT be concluded based on the results in this paper?
    \normalsize   
    \begin{enumerate}[leftmargin=*]
        \item \textbf{\textit{The alignment models struggle to generalize, as they perform better on recipes for dishes seen in training than those not seen in training.}}
        \item \textit{Simply aligning recipe actions based on their sequential order is not a viable baseline for the task, but using cosine similarity works better.}
        \item \textit{Incorporating graphical information about a recipe improves the alignment model’s performance on aligning recipe actions.}
        \item \textit{None of the proposed systems achieve human performance, demonstrating the difficulty of the recipe alignment problem.}
    \end{enumerate}
    \normalsize
    
\end{enumerate}

\subsection{Paper C Questions}
\begin{enumerate}[leftmargin=*]

    \item \textbf{Motivation:} Which of the following is NOT a motivation of this work?
    \normalsize   
    \begin{enumerate}[leftmargin=*]
        \item \textit{Understanding tables requires reasoning over multiple fragments of text in different cells that may not otherwise seem related.}
        \item \textbf{\textit{Tables are uniquely challenging to understand because they convey explicit information that unstructured text does not.}}
        \item \textit{Transformer-based language models have exceeded human performance on a variety of natural language understanding tasks.}
        \item \textit{Semi-structured text can convey unstated information that state-of-the-art language models may fail to recognize.}
    \end{enumerate}
    \normalsize
    
    \item \textbf{Problem Definition:} What task is being studied in this paper?
    \normalsize   
    \begin{enumerate}[leftmargin=*]
        \item \textit{Question answering based on Wikipedia info-boxes.}
        \item \textit{Relation extraction for cells in Wikipedia info-boxes.}
        \item \textit{Table information summarization.}
        \item \textbf{\textit{Textual entailment based on a semi-structured context.}}
    \end{enumerate}
    \normalsize
    
    \item \textbf{Approaches:} What are the inputs to the \textsc{RoBERTa} baseline model with TabFact structured premise representation?
    \normalsize   
    \begin{enumerate}[leftmargin=*]
        \item \textit{A table converted to a paragraph, and a proposed fact about the table.}
        \item \textbf{\textit{A table converted to a set of key-value pairs, and a proposed fact about the table.}}
        \item \textit{A proposed fact about the table, and the most similar sentence to it from the table.}
        \item \textit{A proposed fact about the table, and the most similar three sentences to it from the table.}
    \end{enumerate}
    \normalsize
    
    \item \textbf{Approaches:} What are the outputs of the \textsc{RoBERTa} baseline model with TabFact structured premise representation?
    \normalsize   
    \begin{enumerate}[leftmargin=*]
        \item \textit{A one-sentence summary of the table in text.}
        \item \textit{An inference label of entailment or contradiction.}
        \item \textbf{\textit{An inference label of entailment, neutral, or contradiction.}}
        \item \textit{Tags describing the span of cells that answer the question about the table.}
    \end{enumerate}
    \normalsize
    
    \item \textbf{Implementation:} In which file are tables converted to TabFact structured premises?
    \normalsize   
    \begin{enumerate}[leftmargin=*]
        \item \textbf{\texttt{infotabs-code/scripts/preprocess/ json\_to\_struct.py}}
        \item \texttt{infotabs-code/scripts/preprocess/ json\_to\_wmd.py}
        \item \texttt{infotabs-code/scripts/roberta/ preprocess\_roberta.py}
        \item \texttt{infotabs-code/scripts/roberta/ json\_to\_struct.py}
    \end{enumerate}
    \normalsize
    
    \item \textbf{Results:} What is the difference between the “$\alpha_2$” and “$\alpha_3$” columns in Table 7 of the paper?
    \normalsize   
    \begin{enumerate}[leftmargin=*]
        \item \textit{The $\alpha_2$ test set includes tables from a different domain than the training set, while the $\alpha_3$ test set includes hypothesis sentences that have been adversarially edited by human annotators.}
        \item \textit{The $\alpha_2$ test set includes hypothesis sentences from a different domain than the training set, while the $\alpha_3$ test set includes tables that have been adversarially edited by human annotators.}
        \item \textbf{\textit{Models that overfit to superficial lexical cues will struggle with the $\alpha_2$ test set, while models that overfit to domain-specific statistical cues will struggle with the $\alpha_3$ test set.}}
        \item \textit{Models that overfit to domain-specific statistical cues will struggle with the $\alpha_2$ test set, while models that overfit to superficial lexical cues will struggle with the $\alpha_3$ test set.}
    \end{enumerate}
    \normalsize

    \item \textbf{Conclusion:} Which of the following CANNOT be concluded based on the results in this paper?
    \normalsize   
    \begin{enumerate}[leftmargin=*]
        \item \textit{Pre-trained state-of-the-art natural language inference systems do not perform well when applied directly to tasks requiring reasoning over tables.}
        \item \textbf{\textit{A support vector machine performs better than transformer-based language models on InfoTabs when representing tables as paragraphs.}}
        \item \textit{Encoding a table in structured language rather than an unstructured paragraph helps improve performance of language models on InfoTabs.}
        \item \textit{The proposed systems for InfoTabs tend to struggle most with cross-domain generalization.}
    \end{enumerate}
    \normalsize
    
\end{enumerate}

\end{document}